\documentclass[a4paper,fleqn]{cas-dc}

\usepackage[authoryear]{natbib}

\usepackage[switch]{lineno}
\usepackage{caption}
\usepackage{hyperref}
\usepackage{threeparttable}

\def\tsc#1{\csdef{#1}{\textsc{\lowercase{#1}}\xspace}}
\tsc{WGM}
\tsc{QE}
\tsc{EP}
\tsc{PMS}
\tsc{BEC}
\tsc{DE}
\begin{document}
%\listoffigures
%\listoftables
%\hypersetup{
%    colorlinks=true,
%    linkcolor=blue,
%    filecolor=blue,      
%    urlcolor=blue,
%    citecolor=blue,
%}
\captionsetup[figure]{labelfont={bf},labelformat={default},labelsep=period,name={Fig.}}
\renewcommand*{\figureautorefname}{Fig.}
\let\WriteBookmarks\relax
\def\floatpagepagefraction{1}
\def\textpagefraction{.001}
\shorttitle{Science of The Total Environment}
\shortauthors{Ke Wang et~al.}

\title [mode = title]{A feature-supervised generative adversarial network for environmental monitoring during hazy days}                      
\author[1]{Ke Wang}
\cormark[1]
\ead{kw@cqu.edu.cn}

\author[1]{Siyuan Zhang}
\ead{siyuanzhang@cqu.edu.cn}

\author[2]{Junlan Chen}
\cormark[1]
\ead{junlanchen@cqnu.edu.cn}

\author[3]{Fan Ren}
\ead{renfan@changan.com.cn}

\author[4]{Lei Xiao}
\ead{xiaolei@csrzic.com}

%\fntext[fn1]{These authors contributed equally to this work.}
\cortext[cor1]{Corresponding author.}
\address[1]{School of Automobile Engineering, the Key Lab of Mechanical Transmission, Chongqing University, Chongqing 400044, China}
\address[2]{School of Economics \& Management, Chongqing Normal University, Chongqing 401331, China}
\address[3]{Intelligent Vehicle R$\&$D Institute, Changan Auto Company, Chongqing 401120, China}
\address[4]{CRRC Zhuzhou Institute Co.,Ltd, Zhuzhou, 412001, China} 

\begin{abstract}
The adverse haze weather condition has brought considerable difficulties in vision-based environmental applications. While, until now, most of the existing environmental monitoring studies are under ordinary conditions, and the studies of complex haze weather conditions have been ignored. Thence, this paper proposes a feature-supervised learning network based on generative adversarial networks (GAN) for environmental monitoring during hazy days. Its main idea is to train the model under the supervision of feature maps from the ground truth. Four key technical contributions are made in the paper. First, pairs of hazy and clean images are used as inputs to supervise the encoding process and obtain high-quality feature maps. Second, the basic GAN formulation is modified by introducing perception loss, style loss, and feature regularization loss to generate better results. Third, multi-scale images are applied as the input to enhance the performance of discriminator. Finally, a hazy remote sensing dataset is created for testing our dehazing method and environmental detection. Extensive experimental results show that the proposed method has achieved better performance than current state-of-the-art methods on both synthetic datasets and real-world remote sensing images.
\end{abstract}

%\newpage
%\begin{center}
%\includegraphics[width=6in]{figs/graphicabstract.pdf}
%\end{center}
%\item Using generative adversarial network for environmental monitoring during hazy days
%\item Supervising the model with pairs of hazy and clean image inputs
%\item Introducing several loss functions to constrain the training process
%\item To enhance modeling performance multi-scale discriminator was applied
%\item A hazy remote sensing dataset is created containing synthetic and real hazy images


\maketitle
%\linenumbers
\section{Introduction}

Nowadays, remote sensing technologies such as satellite imagery and unmanned aerial vehicle (UAV) have increasingly been used for environmental monitoring, including wildlife inventorying and monitoring \citep{Korczak-Abshire2019, drones3020039}, fluvial dynamics \citep{mivrijovsky2015multitemporal}, vegetation monitoring \citep{doi:10.1111/ddi.12693, MIRANDA2020135295, von2015deploying, ludovisi2017uav, dandois2010remote}, atmosphere observations \citep{cassano2014observations, witte2017development}. It plays a key role in many vision-based environmental assessments and monitoring systems \citep{RN7668}. However, until now, most of the existing environmental-related researches are under ordinary conditions, while the researches during hazy days have been ignored. There are still some challenges in hazy weather condition pending to be solved: 

(1) in a hazy atmosphere, light emanating from distant sources is often scattered, and the observer can only perceive a reduction in contrast. 

(2) the presence of haze greatly reduces the visibility of outdoor images and affects many advanced environmental monitoring tasks, such as detection and recognition.

Both of these challenges make haze removal a highly needed technique for vision-based environmental systems \citep{RN7312, RN6998, wang2020adaptability}. Single image dehazing methods are mainly based on the atmospheric scattering model, which has been widely used as the description for the hazy image generation process. And the model can be expressed as follow:
\begin{equation}
\begin{split}
	I(x)=J(x)t(x)+A(1-t(x))\label{atmospheric scattering model}
\end{split}
\end{equation}
Where ${I(x)}$ is the hazy image, also the input of the dehazing models, ${J(x)}$ is the hazy-free image, also the output of the dehazing models. $A$ and $t(x)$ represent the global atmospheric light and the medium transmission map, respectively. When atmospheric light is homogeneous, the transmission map can be expressed as follow:
\begin{equation}
\begin{split}
	t(x)={{e}^{-\beta d(x)}}
\end{split}
\end{equation}
Where $\beta $ is defined as the scattering coefficient of the atmosphere, and the $d(x)$ is the scene depth. According to the atmospheric scattering model, we can recover the original hazy-free image via:
\begin{equation}
\begin{split}
	J(x)=\frac{I(x)-A}{t(x)}+A
\end{split}
\end{equation}

In the initial phase of developing haze removal, the prior-based methods are used to estimate parameters of atmospheric scattering models \citep{10.1145/1399504.1360671, 5567108, 6909779, 6751186, 7780554}. For example, \cite{5567108} assumed the value of the dark channel in the clear image is close to zero, then used it to estimate the transmission map. The boundary constraints and context regularization (BCCR) are further enhanced by \cite{6751186} to obtain sharper images. \cite{7128396} developed a color attenuation prior and created a linear model of scene depth for the hazy image, and then learned the model parameters in a supervised way. \cite{7299133} jointly estimated scene depth and recovered the clear latent image from a foggy video sequence. \cite{7780554} proposed a non-local prior, based on the assumption that each color cluster in the clear image becomes a haze-line in RGB space. Despite the extraordinary performance obtained through these methods, it is still easy to violate the adopted priors or assumptions in practice, especially when the scene contains complex or irregular lighting or damage. For example, the assumption proposed by \cite{5567108} does not work well for the scene objects which are similar to the atmospheric light. This usually leads to unsatisfied dehazing quality for sky regions or bright objects.

To overcome the disadvantages of these prior-based methods, recent emphasis has shifted to developing data-driven methods based on deep learning \citep{7539399, 10.1007/978-3-319-46475-6_10, 8237773, 8545522, liu2018cross}. These methods first estimate the transmission map and then use conventional methods to recover clear images, which can avoid inaccurate estimation of physical parameters from a single image. \cite{7539399} proposed a trainable model (DehazeNet) for estimating the transfer matrix from hazy images. \cite{10.1007/978-3-319-46475-6_10} came up with a multi-scale convolutional neural network (MSCNN), which consists of coarse-scale and fine-scale networks to estimate the transmission map. The coarse-scale network estimates the transmission map, which is also improved locally by the ﬁne-scale network. \cite{8237773} proposed an approximation method that absorbs the transmission map and the global atmospheric light coefficient into an intermediate parameter and adopts a neural network to learn it. Generative adversarial network (GAN) is a class of machine learning systems invented by \cite{goodfellow2014generative}. It consists of two neural networks, called generator and discriminator. The generative network generates candidates while the discriminative network evaluates them. The training goal of the generative network is to increase the error rate of the discrimination network. Generally, generators use random input sampled from a predefined latent space (for example, a multivariate normal distribution) as a seed. Thereafter, the discriminator is used to evaluate the candidates generated by the generator. Backpropagation (BP) is applied in both networks so that the generator produces better images, and the discriminator becomes more proficient at labeling synthetic images. Recently, GAN has become a research trend in single image dehazing. \cite{8695091} proposed a multi-tasking method that includes three modules, namely transmission map estimation by GAN, hazy feature extraction and image dehazing. All modules are jointly trained and use image-level loss functions, such as perceptual loss and pixel-wise Euclidean loss. \cite{8578954} developed an end-to-end hazing method based on a conditional generative adversarial network (cGAN) \citep{mirza2014conditional}. Although learning-based methods have made great progress, several factors hinder the performance of these methods, and the results are far from optimal. First, the estimation of transmission map is not always accurate, and some common pre-processing such as guildfiltering or softmatting will further distort the hazy image generation process. Second, existing methods do not consider the possibility of any pair of images have a shared-latent space and do not make full use of the information on clean images to improve the training process. This may hinder the overall dehazing performance.

This study proposes a feature-supervised generative adversarial network for environmental monitoring during hazy days. Its main idea is to train the model under the supervision of feature maps from the ground truth. Specifically, We assume for any given pair of hazy and clean images, there exists a shared-latent space. Based on this assumption, we put this pair of images into two identical encoders, which are part of the generator. Then a feature regularization loss is used to constrain the training process. And for discriminator, multi-scale inputs are applied to improve the performance. Besides, several other loss functions are used to get high-quality hazy-free images, which are not only in style but also in content.

The main contributions of the present study are as follows: (1) Proposing a feature-supervised adversarial generation network that can improve modeling performance. (2) Assuming that any pair of hazy and clean images have similar information in the shared-latent space, which can be used to supervise the training process to get better results. (3) To enhance modeling performance, multi-scale discriminator and several loss functions were applied. (4) A hazy remote sensing dataset with synthetic and real hazy remote sensing images is created for testing our dehazing method and environmental detection. (5) The proposed method can achieve high quality on both synthetic datasets and real-world remote sensing images. 

%----------------------------------------------------------------------------------------------------------------------
\section{Material and methods}
\subsection{Data description} \label{data}

Since it is impractical to obtain paired clean and hazy images of the same view and the same scene at the same time for training, we create a large-scale synthesized dataset for this experiment by \autoref{atmospheric scattering model}. The training set contains indoor and outdoor datasets. The indoor datasets are based on NYU Depth dataset \citep{10.1007/978-3-642-33715-4_54}, where we generate the random atmosphere light $A=\left[ {{m}_{1}},{{m}_{2}},{{m}_{3}} \right]$ with $m\in \left[ 0.7,1.0 \right]$ and select $\beta \in \left[ 0.6,1.8 \right]$. Since the performance of the model heavily depends on the data, we also use outdoor images to increase the diversity of training data. For outdoor datasets, we use $\text{RESIDE-}\beta$ \citep{8451944}, a hazy dataset for image dehazing, which estimates the depth by \cite{7346484} and generates hazy images the same way as indoor datasets. From the indoor and outdoor datasets, we randomly choose 2343 synthesized images as the training set.

And for the test datasets, we construct five kinds of datasets, containing both high-resolution and low-resolution, synthetic and real hazy images, as listed below:

Test Dataset A: Test dataset A consists of 252 synthesized images from the rest of indoor and outdoor datasets. We use this dataset to verify the performance of the proposed method among current state-of-art methods.

Test Dataset B: Test dataset B contains 640 synthesized hazy remote sensing images based on DOTA \citep{Xia_2018}, a large-scale dataset for object detection in aerial images. Those images are from Google Earth, GF-2 or JL-1 with the size in the range from about 800x800 to 4000x4000 pixels. Considering the scene depth of remote sensing images is almost constant, we directly select $t(x)\in \left[ 0.\text{2},\text{0}\text{.6} \right]$ and generate the random atmosphere light $A=\left[ {{m}_{1}},{{m}_{2}},{{m}_{3}} \right]$ with $m\in \left[ 0.7,1.0 \right]$ to create paired hazy remote sensing images.

Test Dataset C: Test dataset C consists of real hazy remote sensing images from {Landsat 8 Operational Land Imager} making use of the bands (2), (3) and (4) as BGR true color, including the forest area, the ocean area and the barren mountain area, which are taken in the eastern coastal and western mountainous areas of China.

Test Dataset D: Test dataset D was made up of real UAV images obtained under hazy weather conditions.

Test Dataset E: Test dataset E contains 289 synthesized hazy remote sensing images based on DOTA with 15 common object categories annotations, including plane, ship, storage tank (ST), baseball diamond (BD), tennis court (TC), basketball court (BC), ground track field (GTF), harbor, bridge, large vehicle (LV), small vehicle (SV), helicopter (HC), roundabout (RA), soccer ball field (SBF) and swimming pool (SP).  To generate hazy images, we use the same way as Test Dataset B. {The main characteristics of the five test dataset are shown in \autoref{detail}.}

\begin{table*}[width=2.1\linewidth,cols=6,pos=h]
\caption{The main characteristics of the five test dataset.}\label{detail}
\begin{tabular*}{\tblwidth}{@{} LLLLLL@{} }
\toprule
  & Test Dataset A & Test Dataset B & Test Dataset C & Test Dataset D & Test Dataset E\\
\midrule
Number               & 252 & 640  & 30 & 120 & 289\\

Synthetic/Real & Synthetic & Synthetic & Real & Real & Synthetic\\
Source               & Ground camera & GF-2/JL-1    & Landsat 8 OLI & UAV &GF-2/JL-1 \\
Size    & $256\times256$ & $800\times800$ to $4000\times4000$ & $600\times600$ to $2000\times2000$ & $1920\times1080$ & $800\times800$ to $4000\times4000$\\

\bottomrule
\end{tabular*}
\end{table*}

%----------------------------------------------------------------------------------------------------------------------
\begin{figure*}
	\centering
		\includegraphics[width=7in]{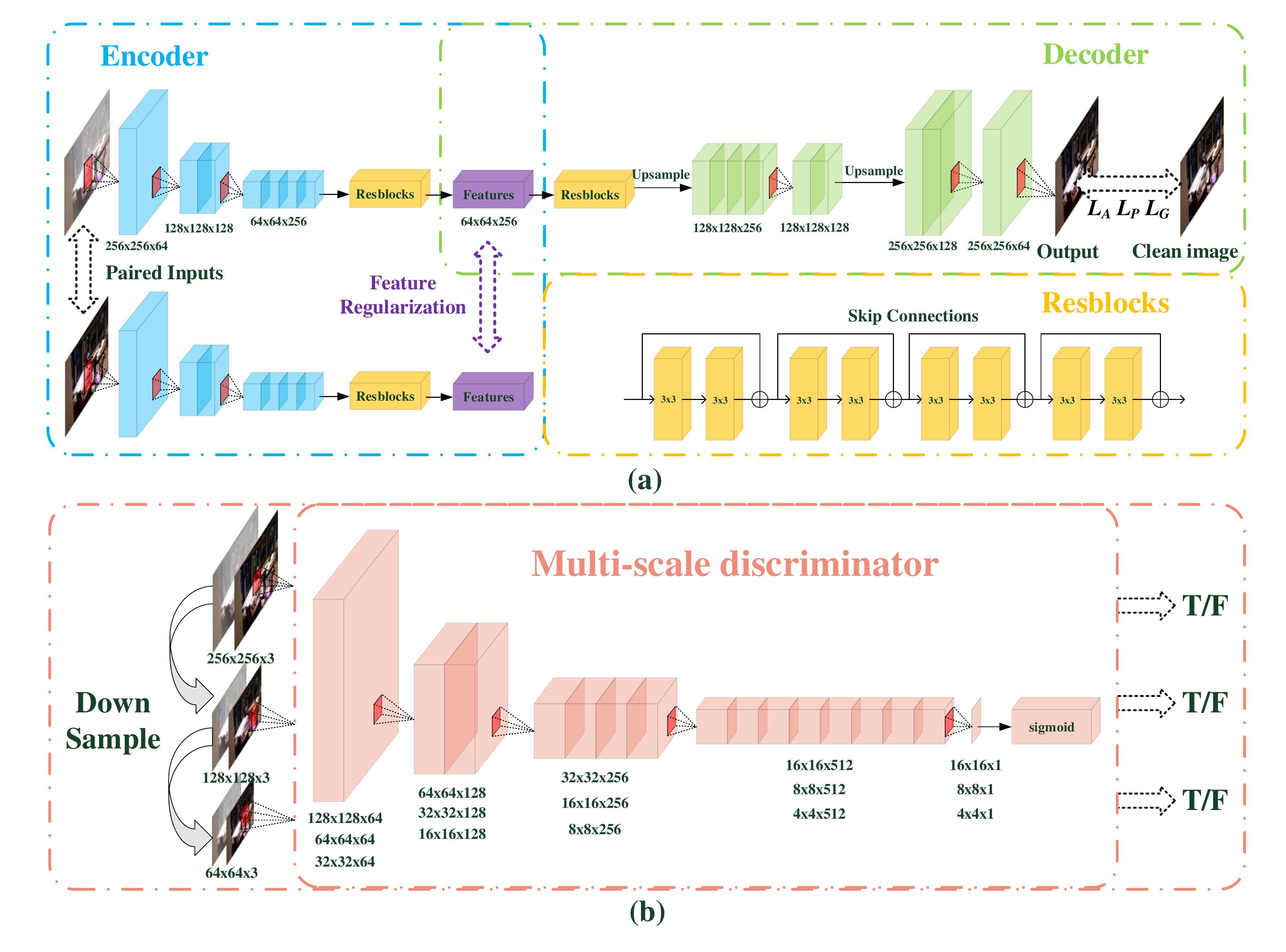}
    
	\caption{The network structure of the proposed method. The generator network contains an encoder and decoder process. The red rectangles denote the convolutional kernel. The hierarchy structures in (b) correspond to the three scales of input respectively.}
	\label{framework}
\end{figure*}

\begin{table}[width=.9\linewidth,cols6,pos=h]
\begin{threeparttable}
\caption{Architecture of generator and parameter setting.}\label{Generator}
\begin{tabular*}{\tblwidth}{@{} LLLLLL@{} }
\toprule
  & Layer & Channel & Kernel Size & Stride & Pad\\
\midrule
Encoder & 1 & 64  & $7\times7$ & 1 & $3\times3$\\
        & 2 & 128 & $4\times4$ & 2 & $1\times1$\\
        & 3 & 256 & $4\times4$ & 2 & $1\times1$\\
        & 4-11 & 256 & $3\times3$ & 1 & $1\times1$\\
Decoder & 1-8 & 256   & $3\times3$ & 1 &$1\times1$ \\
        & 9 & 128 & $5\times5$ & 1 & $2\times2$\\
		& 10 & 64 & $5\times5$ & 1 & $2\times2$\\
		& 11 & 3 & $7\times7$ & 1 & $3\times3$\\
\bottomrule
\end{tabular*}
\begin{tablenotes}
\item[1] There are two upsampling operations with 2 scales between layers 8, 9 and 9, 10.
\end{tablenotes}
\end{threeparttable}
\end{table}

\begin{table}[width=.9\linewidth,cols=5,pos=h]
\caption{Architecture of discriminator and parameter setting.}\label{Discriminator}
\begin{tabular*}{\tblwidth}{@{} LLLLL@{} }
\toprule
 Layer & Channel & Kernel Size & Stride & Pad\\
\midrule
1 & 64  & $4\times4$ & 2 & $1\times1$\\
2 & 128 & $4\times4$ & 2 & $1\times1$\\
3 & 256 & $4\times4$ & 2 & $1\times1$\\
4 & 512 & $4\times4$ & 2 & $1\times1$\\
5 & 1   & $1\times1$ & 1 & \\
\bottomrule
\end{tabular*}
\end{table}

\subsection{Model}
\subsubsection{Architecture of proposed method}
It can be observed from \autoref{atmospheric scattering model} that there exist two important parameters in the dehazing process, which are accurately estimating transmission map and atmospheric light. Between these, the transmission map is known as the key to achieving haze removal \citep{8578435, AAAI1817047}. For estimating these two parameters better and getting hazy-free images directly, we proposed a feature-supervised model based on GAN that the generator is composed of an encoder and a decoder, and the discriminator is a convolutional neural network.

In our model, the function of the generator is to get a clear image from a hazy input image. Therefore, it should not only preserve the structure and detail information of the input image but also remove the haze as much as possible. Different from the traditional GAN, we assume that any pair of hazy and clean images have similar information in the shared-latent space, which can be used to supervise the training process to get high-quality feature maps. In addition, learning more robust and abstract hierarchical features will help to improve the representation learning performance of the training model. Based on these, we feed the pair of hazy and clean images into the encoder to obtain two similar feature maps. Then use a loss function called feature regularization loss to ensure the feature map from the hazy image remains more useful information for the decoding process. During the encoding process, the feature map from the clean image contains priors of the hazy-free image, therefore they can be used to supervise the network to produce better results. In order to get more information from feature maps, we use Resblocks before getting the feature maps to concatenate the channels of the symmetric layers. At the same time, we use Resblocks to break through the bottleneck of information in the decoder. Furthermore, we introduce a new loss function including adversarial loss, style loss, feature regularization loss and perceptual loss to constrain the training process, which will be introduced in detail in the next part. 

As shown in \autoref{framework}, the generator contains an encoding and decoding process. The encoding process is mainly based on convolutional (\textit{Conv}) operations and provides feature maps to the symmetric layer of the decoding process. The decoding process mainly uses upsampling and convolutional operations. The details of proposed generator structures and parameter settings are depicted in \autoref{Generator}. A hazy image is fed into the generator as input, then convolutional (encoder) and upsample (decoder) layers are designed to extract features and construct hazy-free images, respectively. In particular, the encoder contains one $7\times7$ \textit{Conv} layer with stride-1 and $3\times3$ padding, two $4\times4$ \textit{Conv} layers with stride-2 and $1\times1$ padding and one Resblocks including eight $1\times1$ \textit{Conv} layers with stride-1 and $1\times1$ padding. The decoder contains one Resblocks including eight $3\times3$ \textit{Conv} layers with stride-1 and $1\times1$ padding, two $5\times5$ \textit{Conv} layers with stride-1 and $2\times2$ padding, one $7\times7$ \textit{Conv} layer with stride-1 and $3\times3$ padding and two upsampling operations with 2 scales between layers 8, 9 and 9, 10. During this process, we use the proposed loss functions to constrain it for high-quality results. The benefits of supervised learning and loss functions will be discussed in \autoref{effect of the loss}. 

The discriminator is used to distinguish whether an image is real or not. In our model, the hazy-free images from generator should be fake for discriminator, and the ground truth is real. The goal of generator is to generate an image that can fool the discriminator and make the discriminator thinks it is real. Actually, it should be fake as we just mentioned. But for discriminator, it is designed to distinguish whether the image comes from the generator or the ground truth and keep from being fooled as much as possible. The game between the generator and discriminator improves the performance of our model. In order to generate a high-resolution image for environmental monitoring, it is necessary to increase the receptive field of the discriminator. Furthermore, considering simply increases the number and complexity of network may lead to overfitting (the accuracy of training is high and the testing is low) and challenges for training, we adopt multi-scale discriminators which consist of three discriminators (D1, D2 and D3) that discriminate different resolution versions of the images from generator and training datasets. These three discriminators have the same network structure, and the only difference among D1, D2 and D3 is the different sizes of their input images. As shown in \autoref{framework}, we downsample the target clean image and generative image by average pooling to create an image pyramid with 2 scales. Therefore, the inputs of three discriminators are with the shape of $256\times256\times3$, $128\times128\times3$, $64\times64\times3$, respectively.

The architecture of the proposed discriminator and parameter setting are depicted in \autoref{Discriminator}. The basic operations of discriminator are convolutional and downsample. In particular, the discriminator contains four $3\times3$ \textit{Conv} layers with stride-2 and $1\times1$ padding and one $1\times1$ \textit{Conv} layer with stride-1 and no padding. For the final layer of the discriminator, we apply a sigmoid function to the feature maps in order to normalize the probability score into [0,1], where 0 and 1 represent fake and real, respectively. 

\subsubsection{Loss function}
Our objective function contains four terms: an adversarial loss, a style loss, a feature regularization loss and a perceptual loss.

\textbf{Adversarial loss:} Let us denote $\{{{I}_{i}},i=1,2,...,N\}$ and $\{{{J}_{i}},i=1,2,...,N\}$ are two sets of training samples corresponding to hazy images and ground truth (clean images), respectively. In order to generate high-quality hazy-free images, we train the multi-scale discriminator D to detect whether an image is real or fake. Besides, the generator based on variational autoencoder (VAE) is adversarially trained to “fool” the discriminator. Which can be expressed as:
\begin{equation}
\begin{split}
	{{L}_{A}}=\frac{1}{N}\sum\limits_{i=1}^{N}{\log (1-D({{I}_{i}},{{{\tilde{J}}}_{i}}))}\label{adversarial loss}
\end{split}
\end{equation}
Where ${{\tilde{J}}_{i}}$ is the output of the generator G, also can be denoted as ${{\tilde{J}}_{i}}=G\left( {{I}_{i}} \right)$. However, sometimes GANs are unstable to train, resulting in artifacts in output images. Moreover, we found the variational autoencoder-generative adversarial network (VAE-GAN) algorithm using this function is not able to remove the haze well and will generate some artifacts and color distortions on output images due to its disadvantages. Also, we found that cycle-consistent adversarial networks (CycleGAN), a useful style translation network based on GAN, it still has the problem like VAE-GAN. As shown in the next section, both the visual results and the quantitative results indicate that only using \autoref{adversarial loss} does not generate satisfied hazy-free images.

\textbf{Perceptual loss:} In order to generate more realistic images, we use the perceptual loss based on the pre-trained VGG (very deep convolutional networks proposed by Visual Geometry Group) feature, which is defined as:
\begin{equation}
\begin{split}
	{{L}_{p}}=\frac{1}{{{C}_{k}}{{H}_{k}}{{W}_{k}}}\left\| {{\phi }_{k}}({{{\tilde{J}}}_{i}})-{{\phi }_{k}}({{J}_{i}}) \right\|_{2}^{2}
\end{split}
\end{equation}
Where ${{\phi }_{k}}$ is the feature maps in the $k-th$ layer of the VGG network \citep{simonyan2014very}, which is re-trained on ImageNet \citep{Russakovsky2015}. ${{C}_{k}},{{H}_{k}},{{W}_{k}}$ are the dimensions of the feature maps. Perceptual loss encourages the output images and the target images to have similar feature representations rather than encouraging them to exactly match the pixels. That is because in the early layers, minimizing the perceptual loss tends to produce images that are visually indistinguishable from the target. And in high layers, image content and overall spatial structure are preserved but color, texture and exact shape are not. In our experiments, we found that perceptual loss can help the details restoration and haze removal but it also generates artifacts.

\textbf{Style loss:} In order to solve the problem that the differences in the style, such as colors, textures common patterns, etc., we introduce the style loss, which defines as the difference between the Gram matrices of the output and the target images:
\begin{equation}
\begin{split}
	{{G}_{k}}=\frac{1}{{{C}_{k}}{{H}_{k}}{{W}_{k}}}\sum\limits_{h=1}^{{{H}_{k}}}{\sum\limits_{w=1}^{{{W}_{k}}}{{{\phi }_{k}}}}{{(x)}_{h,w,c}}{{\phi }_{k}}{{(x)}_{h,w,{{c}^{,}}}}
\end{split}
\end{equation}
The gram matrix can be computed by reshape ${{\phi }_{k}}\left( x \right)$ into a matrix $\psi $ of shape ${{C}_{k}}\times {{H}_{k}}{{W}_{k}}$, then ${{G}_{k}}=\psi {{\psi }^{T}}/{{C}_{k}}{{H}_{k}}{{W}_{k}}$. Then we compute the squared Frobenius norm of the difference between the Gram matrices of the output and corresponding clean images:
\begin{equation}
\begin{split}
	{{L}_{S}}=\left\| {{G}_{k}}({{{\tilde{J}}}_{i}})-{{G}_{k}}({{J}_{i}}) \right\|_{F}^{2}
\end{split}
\end{equation}

We found out that minimize the style loss preserves stylistic features from the target images can get better results than only use perceptual loss. However, the hazy-free images still have some hazy zone and artifacts.

\textbf{Feature regularization loss:} Considering that any pair of hazy and clean images may have the same space when we encode them, we introduce the feature regularization loss to get more information and constrains from the corresponding clean images. The loss function is calculated as:
\begin{equation}
\begin{split}
	{{L}_{FR}}=\frac{1}{N}\sum\limits_{i=1}^{N}{\left( {{\left\| {{E}_{k}}({{I}_{i}})-{{E}_{k}}({{J}_{i}}) \right\|}_{1}} \right)}
\end{split}
\end{equation}
Where $E$ and $k$ represent the encoder and the $k-th$ Conv layer in the encoder, respectively.

Finally, we combine the adversarial loss, perceptual loss, style loss, feature regularization loss to regularize the proposed generative network, which is defined as:
\begin{equation}
\begin{split}
	L=\gamma_1 {{L}_{A}}+\gamma_2 {{L}_{P}}+\gamma_3 {{L}_{S}}+\gamma_4 {{L}_{FR}}\label{loss function}
\end{split}
\end{equation}
where $\gamma_1 ,\gamma_2 ,\gamma_3 ,\gamma_4 $ are the positive weights. The generator is trained to minimize \autoref{loss function}.

After obtaining the intermediate generator G, we update the discriminator D by:
\begin{equation}
\begin{split}
	\underset{D}{\mathop{\max }}\,\frac{1}{N}\sum\limits_{m=1}^{3}{\sum\limits_{i=1}^{N}{\left( \log \left( 1-{{D}_{m}}\left( {{I}_{i}},{{{\tilde{J}}}_{i}} \right) \right) \right)}}+\log \left( {{D}_{m}}\left( {{I}_{i}},{{J}_{i}} \right) \right)
\end{split}
\end{equation}

\subsection{Performance evaluation}
\subsubsection{Performance evaluation for hazy-free images}
In order to get higher performance in environmental detection, we need to make sure the hazy-free images have higher quality. Based on this, we evaluate our algorithm on the synthetic dataset and compare it with several state-of-the-art single image dehazing methods using Peak Signal to Noise Ratio (PSNR) and Structural Similarity Index (SSIM).

PSNR is an engineering term for the ratio between the maximum possible power of a signal and the power of corrupting noise that affects the fidelity of it is representation. It measures the similarity between two images (how two images are close to each other). In our model, the reference images is ${J}$ and the test image is ${\tilde{J}}$, both of size $U\times V$, the PSNR is defined as follows:
\begin{equation}
\begin{split}
	PSNR\text{=10}{{\log }_{10}}\left( {{P}^{\text{2}}}/MSE \right)
	\end{split}
\end{equation}
\begin{equation}
\begin{split}
	MSE=\frac{1}{UV}{{\sum\limits_{u=1}^{U}{\sum\limits_{v=1}^{V}{\left( {{J}_{uv}}-{{{\tilde{J}}}_{uv}} \right)}}}^{\text{2}}}
\end{split}
\end{equation}
Where 
${{J}_{uv}}$ is the ${{u}^{th}}$ row and the ${{v}^{th}}$ column pixel in the reference image,
${{\tilde{J}}_{uv}}$ is the ${{u}^{th}}$ row and the ${{v}^{th}}$ column pixel in the reference image,
$P$ is the dynamic range of pixel values, or the maximum value that a pixel can take (equals to 255 for 8-bit images).

The PSNR value approaches infinity as the MSE approaches zero; this shows that a higher PSNR value provides a higher image quality. At the other end of the scale, a small value of PSNR implies high numerical differences between images.

The SSIM is a metric used to measure the similarity between two images and is considered to be correlated with the quality of perception of the human visual system (HVS). Instead of using the traditional error summation methods, the SSIM is designed by modeling any image distortion as a combination of three factors that are loss of correlation, luminance distortion and contrast distortion. The SSIM is defined as follows:
\begin{equation}
\begin{split}
\begin{aligned}
&SSIM\text{=}l\left( \tilde{J},J \right)c\left( \tilde{J},J \right)s\left( \tilde{J},J \right) \\ 
&l\left( \tilde{J},J \right)=\frac{2{{\mu }_{J}}{{\mu }_{{\tilde{J}}}}+{{C}_{1}}}{\mu _{J}^{2}+\mu _{{\tilde{J}}}^{2}+{{C}_{1}}} \\ 
&c\left( \tilde{J},J \right)=\frac{2{{\sigma }_{J}}{{\sigma }_{{\tilde{J}}}}+{{C}_{2}}}{\sigma _{J}^{2}+\sigma _{{\tilde{J}}}^{2}+{{C}_{2}}} \\ 
&s\left( \tilde{J},J \right)=\frac{{{\sigma }_{J\tilde{J}}}+{{C}_{3}}}{{{\sigma }_{J}}{{\sigma }_{{\tilde{J}}}}+{{C}_{3}}} \\ 
\end{aligned}
\end{split}
\end{equation}
Where ${{\mu }_{J}},{{\mu }_{{\tilde{J}}}}$ denote the mean values of reference and test images, ${{\sigma }_{J}},{{\sigma }_{{\tilde{J}}}}$ denote the standard deviation of reference and test images, and ${{\sigma }_{J\tilde{J}}}$ is the covariance of both images. The positive constants ${{C}_{1}},{{C}_{1}},{{C}_{3}}$ are used to avoid a null denominator.
$l\left( \tilde{J},J \right)$ is the luminance comparison function which measures the closeness of the two images’ mean luminance. This factor is maximal and equal to 1 only if ${{\mu }_{J}}={{\mu }_{{\tilde{J}}}}$. $c\left( \tilde{J},J \right)$ is the contrast of the two images and is maximal and equal to 1 only if ${{\sigma }_{J}}={{\sigma }_{{\tilde{J}}}}$. $s\left( \tilde{J},J \right)$ is the structure comparison function which measures the correlation coefficient between the two images. The positives value of the SSIM index are in $\left[ \text{0},\text{1} \right]$ . A value of 0 means no correlation between images, and 1 means these two images are exactly the same.

\subsubsection{Performance evaluation for object detection}
For object detection, we use average precision (AP) and mean average precision (mAP). First, let us introduce the recall and precision. Precision measures how accuracy is the predictions and recall measures how good the algorithm to find all the positives. They are defined as follows:
\begin{equation}
\begin{split}
\begin{aligned}
& Precision=\frac{TP}{TP+FP} \\ 
 & Recall=\frac{TP}{TP+FN} \\ 
\end{aligned}
\end{split}
\end{equation}
Where $TP,FP,FN$ are denote true positive, false positive, false positive.The AP computes the average precision value for recall value over 0 to 1, which is defined as follows:
\begin{equation}
\begin{split}
\begin{aligned}
AP\text{=}\int_{\text{0}}^{\text{1}}{p\left( r \right)dr}
\end{aligned}
\end{split}
\end{equation}
Where $p\left( r \right)$ represents the precision-recall curve with recall as the x-axis and precision as the y-axis.
The mAP for object detection is the average of the AP calculated for all the categories. It can be expressed as
\begin{equation}
\begin{split}
\begin{aligned}
mAP=\frac{1}{Q}\sum\limits_{q=1}^{Q}{A{{P}_{q}}}
\end{aligned}
\end{split}
\end{equation}

\subsection{Training settings}
We trained the proposed method with the following settings. Weighting factors $\gamma_1 \text{=1}$, $\gamma_2 \text{=1}$, $\gamma_3 \text{=50}$ and $\gamma_4 \text{=0}\text{.01}$. In this way, parameters in style loss will be more important than others and parameters in feature regularization will be less important. The Adam (adaptive moment estimation) optimization algorithm is used during training, with the learning rate of 0.0001 and weight decay of 0.001. The learning rate is updated by multiplying gamma which we set to 0.5 with the step size of 5000 during training, and our training process is stopped at 300,000 iterations. The proposed algorithm is implemented in Pytorch on a computer with an Nvidia Titan-XP GPU.

%----------------------------------------------------------------------------------------------------------------------
\begin{figure*}
	\centering
		\includegraphics[width=7in]{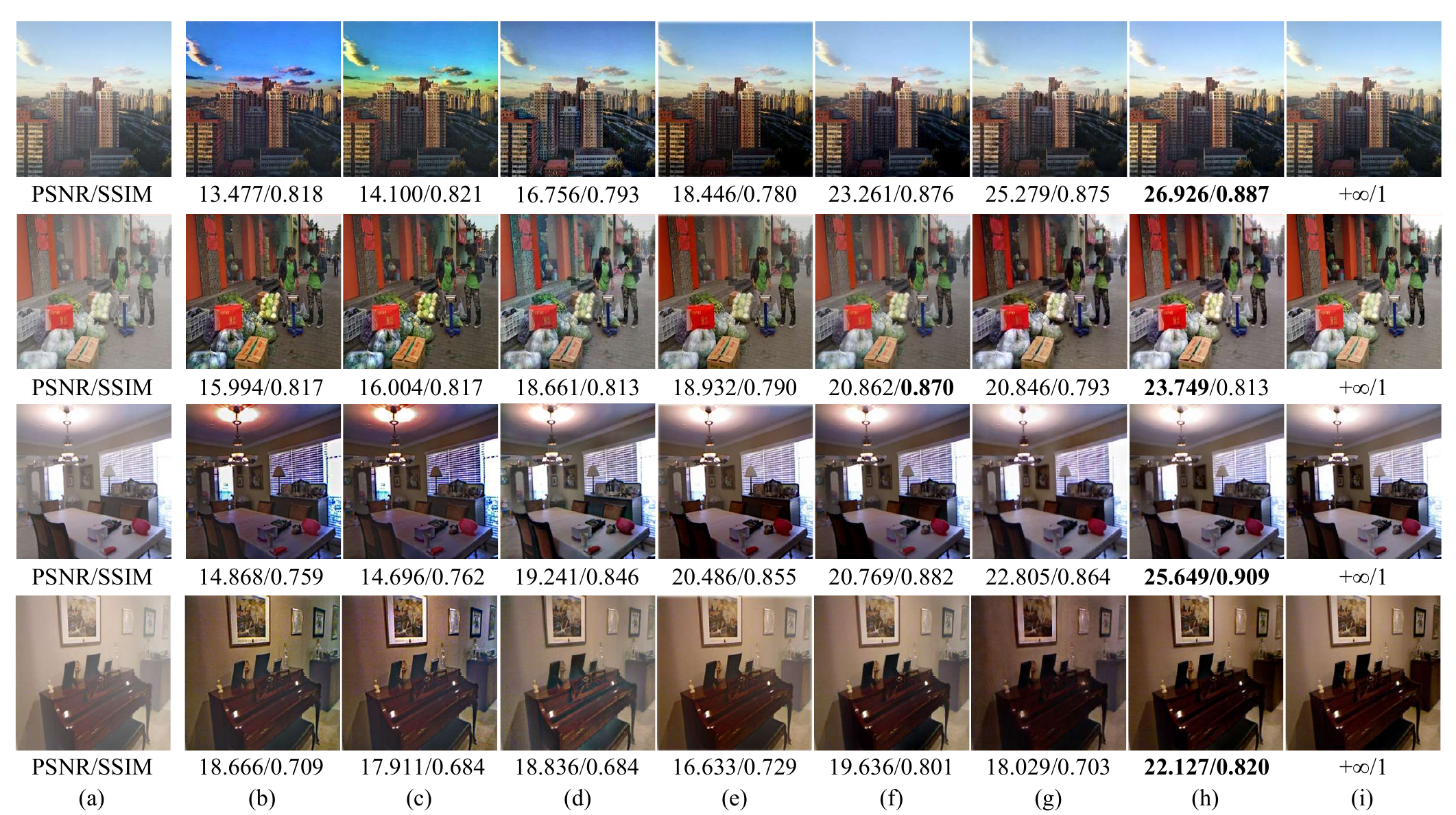}
    
	\caption{Dehazed results on Test Dataset A. Below each picture are the corresponding PSNR and SSIM values with the highest value of each input hazy image bolded. From left to right:(a)input (b)DCP (c)BCCR (d)AMEF (e)AOD-Net (f)DehazeNet (g)CycleGAN (h)Ours (i)Ground truth.}
	\label{resultA}
\end{figure*}

\section{Results and discussion}
In this section, we evaluate our algorithm on five testing datasets to illustrate the efficiency of our method and the application in environmental monitoring. We compare the proposed method with six state-of-art methods of dark channel prior (DCP) \citep{5567108}, BCCR \citep{6751186}, artificial multiple exposure fusion (AMEF) \citep{GALDRAN2018135}, all-in-one dehazing network (AOD-Net) \citep{8237773}, DehazeNet \citep{7539399}, and CycleGAN \citep{8237506}.

\subsection{Quantitative evaluation and comparison on Test Dataset A}
To illustrate the efficiency of our model, we evaluate it on synthetic dataset and compare it with some other state-of-the-art methods using PSNR and SSIM. \autoref{Result} shows the average PSNR and SSIM (PSNR\_AVG and SSIM\_AVG) results on Test Dataset A. The proposed method generates higher PSNR than other algorithms due to adversarial learning and proposed loss function, although the SSIM shows the second best results. According to the table, our method outperforms the others at least 2.8\% in terms of PSNR. Furthermore, we calculate the standard deviation of PSNR and SSIM (PSNR\_SD and SSIM\_SD). The results of standard deviation show that our method is more stable generally.

\autoref{resultA} displays four examples from the Test Dataset A. It is clear to see that the results of DCP have some strong color distortions, especially when the targets are similar to the atmosphere light. The reason for such a phenomenon mainly put down to the inaccurate estimation of the transmission map. Although BCCR  has improved compared to DCP, the color distortions are still the reason that the results look unreal. For AMEF, which removing haze without relying on the inversion of a physical model of hazy formation, the distortions are lighter than DCP and BCCR, but it still has some. For AOD-Net, DehazeNet and CycleGAN, although it generates better results than prior-based method by using CNN to estimate the transmission map, the results still have some hazy residuals in the estimated images. This is mainly due to the underestimation of the hazy level. In contrast, the proposed method generates much cleaner images with fewer hazy residuals and artifacts due to the feature-supervised adversarial learning and proposed loss functions. In addition, our method is the closest one to ground truth, which can be reflected in the PSNR and SSIM below.

\begin{table*}[width=2.1\linewidth,cols=8,pos=h]
\caption{The average and standard deviation of PSNR and SSIM in Test Dataset A.}\label{Result}
\begin{tabular*}{\tblwidth}{@{} LLLLLLLL@{} }
\toprule
Metrics & DCP & BCCR & AMEF & AOD-Net & DehazeNet & CycleGAN & Our\\
\midrule
PSNR\_AVG & 15.880  & 16.049 & 18.559 & 19.357 & 22.573 & 21.892 & 23.215\\
SSIM\_AVG & 0.796   & 0.795  & 0.808  & 0.834  & 0.880  & 0.824  & 0.849\\
PSNR\_SD  & 2.501   & 2.415  & 2.254  & 2.206  & 3.083  & 2.803  & 2.530\\
SSIM\_SD  & 0.055   & 0.058  & 0.075  & 0.062  & 0.060  & 0.057  & 0.042\\
\bottomrule
\end{tabular*}
\end{table*}

\begin{table*}[width=2.1\linewidth,cols=4,pos=h]
\caption{Quantitatively evaluate the effect of the different loss functions on Test Dtaset A.}\label{Resultloss}
\begin{tabular*}{\tblwidth}{@{} LLLL@{} }
\toprule
Metrics & ${L}_{A}+{L}_{P}$& ${L}_{A}+{L}_{P}+{L}_{FR}$ &${L}_{A}+{L}_{P}+{L}_{FR}+{L}_{S}$\\
\midrule
PSNR\_AVG & 22.708& 22.924& 23.215\\
SSIM\_AVG & 0.842& 0.840& 0.849\\
PSNR\_SD  & 2.714& 2.699& 2.530\\
SSIM\_SD  & 0.045& 0.046& 0.042\\
\bottomrule
\end{tabular*}
\end{table*}

\subsection{Effect of loss functions} \label{effect of the loss}
We introduce some loss functions to generate high-quality defogging images. To assess the impact of the loss function, we show the impact of each item in \autoref{Resultloss}. The results are conducted on Test Datasets A with the same settings.

The results from the first column and the second column show that ${L}_{FR}$ helps to improve the PSNR value by contacting feature maps from hazy images and clear images. That is because using feature regularization loss helps to get more information from the shared-latent space so that the decoder can generate robust and abstract hierarchical feature maps. Besides, we noticed that due to the use of the gram matrix, the PSNR and SSIM values generated by the method with ${L}_{S}$ are the highest compared to the first two combinations. That is because the gram matrix solves the problem that the differences in the style, such as colors, textures common patterns and helps the model to find and maintain the style of normal scenes.

\begin{figure*}
	\centering
		\includegraphics[width=7in]{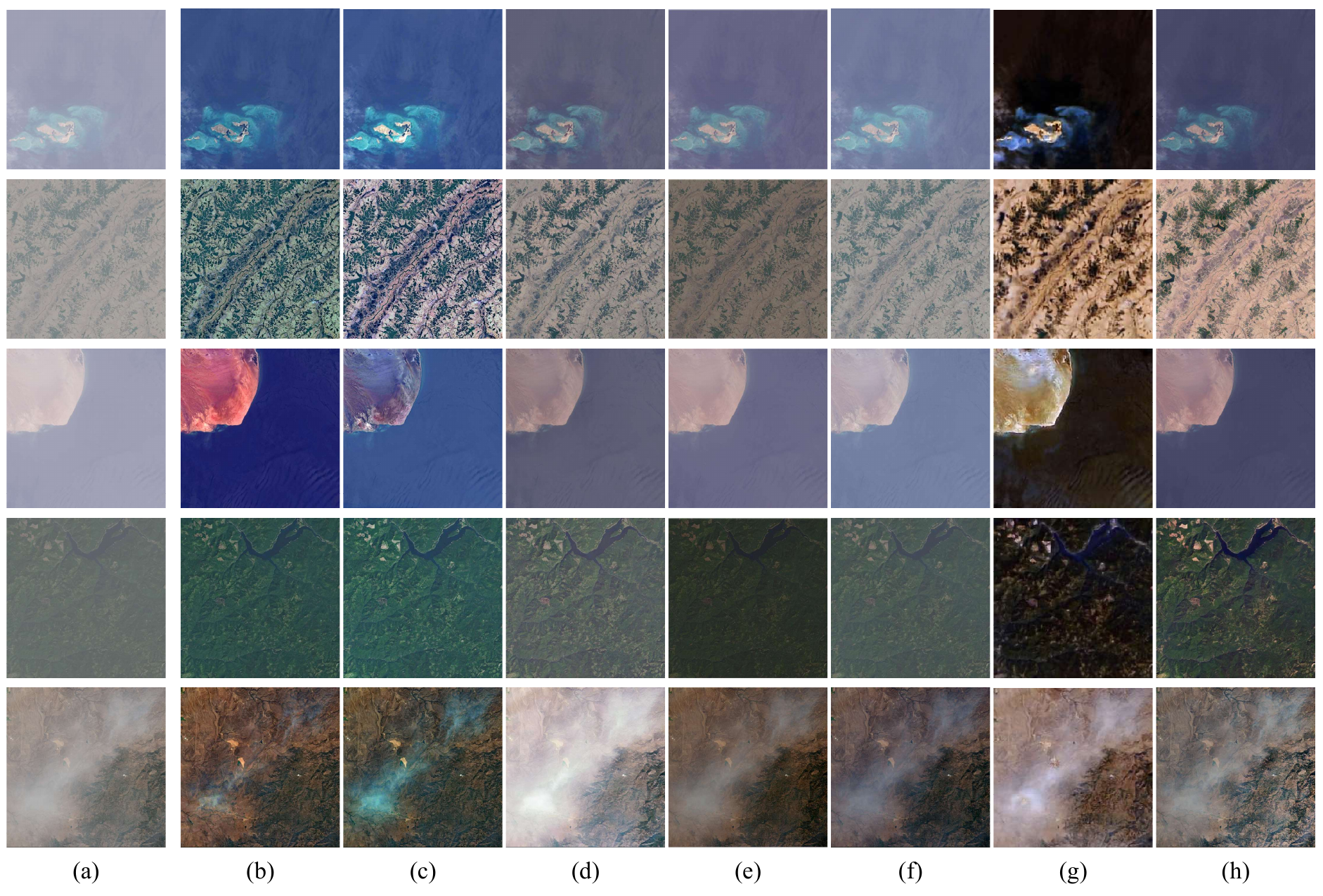}
    
	\caption{Dehazed results on Test Dataset C. From left to right:(a)input (b)DCP (c)BCCR (d)AMEF (e)AOD-Net (f)DehazeNet (g)CycleGAN (h)Ours.}
	\label{landsat}
\end{figure*}

\begin{figure*}
	\centering
		\includegraphics[width=7in]{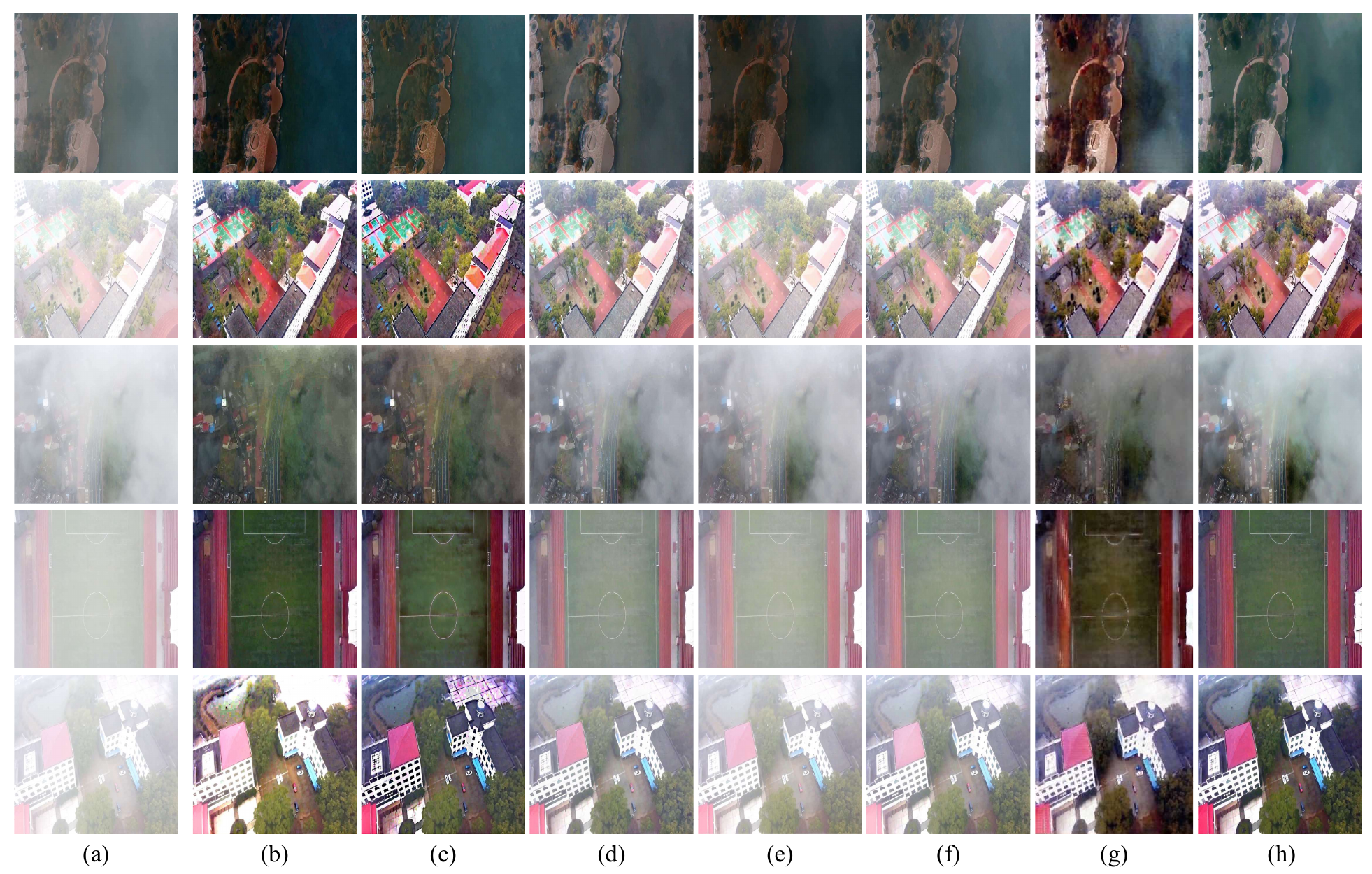}
    
	\caption{Dehazed results on Test Dataset D. From left to right:(a)input (b)DCP (c)BCCR (d)AMEF (e)AOD-Net (f)DehazeNet (g)CycleGAN (h)Ours.}
	\label{UAV}
\end{figure*}

\begin{table}[width=.9\linewidth,cols=4,pos=h]
\caption{The average and standard deviation of PSNR and SSIM on different pixel ranges of Test Dataset B.}\label{result pixel}
\begin{tabular*}{\tblwidth}{@{} LLLLLLLL@{} }
\toprule
Metrics & TDB1 & TDB2 &TDB3\\
\midrule
PSNR\_AVG & 21.649& 21.881& 21.968\\
SSIM\_AVG & 0.862& 0.869& 0.877\\
PSNR\_SD  & 3.016& 3.199& 2.960\\
SSIM\_SD  & 0.072& 0.074& 0.077\\
\bottomrule
\end{tabular*}
\end{table}
\subsection{Quantitative evaluation on different pixel ranges of Test Dataset B}
Since remote sensing images are commonly composed of thousands to tens of thousands of pixels in lines, and the image’s size of Test Dataset A is 256x256, we divide test dataset B into three parts to evaluate the performance on different pixel ranges. The first part (TDB1) consisted of 270 images with total pixels less than $\text{1}{{\text{0}}^{\text{6}}}$. The second part (TDB2) consisted of 261 images with total pixels between $\text{1}{{\text{0}}^{\text{6}}}$ and $\text{2}\times {{10}^{6}}$. And the last part (TDB3) consisted of 109 images with total pixels beyond $\text{2}\times {{10}^{6}}$. From \autoref{result pixel}, we can find that no matter the average or standard deviation, there is almost no difference between PSNR and SSIM in different pixel ranges which shows our method is applicable to remote sensing images.

\subsection{Qualitative evaluation and comparison on Test Dataset C}
To further evaluate the performance of proposed method on real-world remote sensing images rather than synthetic hazy images, we obtain real hazy remote sensing images from {Landsat 8 Operational Land Imager} making use of the bands (2), (3) and (4) as BGR true color and compare the results with other state-of-art dehazing algorithms. A satisfactory result is that our model also shows higher performance on real-world remote sensing images. As revealed in \autoref{landsat}, the traditional algorithms, like DCP, BCCR and AMEF generate unrealistic tones which cause the images color-distorted just like the results on synthetic datasets, especially the DCP and BCCR, which have strong color distortions. For AMEF, due to the artificial multiple-exposure image fusion, it can get more realistic results than the other two traditional methods. However, it still remains some hazy regions. The deep learning methods, AOD-Net and DehazeNet use a CNN to estimate the transmission map and then use conventional method to recover clear images. However, the results still contain some hazy residuals and color distortions. Especially the AOD-Net, the color distortions are much stronger than DehazeNet in some cases, such as the last two images. And for CycleGAN, there exist strong color distortions in hazy-free images. Furthermore, it generates some artifacts, which is one of the problems of GAN. Different from all those methods, the proposed algorithm is based on the supervised by paired clear images, which helps us to estimate the transmission map and the atmospheric light precisely. Therefore, the results generated by our methods are much clear than any other algorithms as shown in \autoref{landsat}. specifically, for the last non-uniform hazy image, all the methods are failed to remove all the haze, that is because our training dataset only includes the uniform hazy images.

\subsection{Qualitative evaluation and comparison on Test Dataset D}
The dehazing results obtained with the real hazy UAV images are shown in \autoref{UAV}. For the traditional methods, DCP and BCCR can remove most of the haze even the challenge one (the third image), but still tend to over-enhance the image and cause strong color distortions. Although like the previous experiments, AMEF can control the color distortion better than DCP and BCCR, it still has some hazy regions, which is similar to AOD-Net and DehzeNet. The difference is that in some cases, the AOD-Net has stronger color distortions than AMEF and DehazeNet, such as the first image. And for CycleGAN, it still exists the problem of instability like the original GAN, which causes the unsatisfactory results on both real hazy remote sensing and UAV images. In contrast, although the developed algorithm can’t remove the haze of the third challenge image effectively like others, it obtains the most pleasing visual results on real UAV images, which demonstrates the effectiveness of the proposed method.

\begin{table*}[width=2.1\linewidth,cols=17,pos=h]
\begin{threeparttable}
\caption{The quantitative detection results on hazy-free images (HFI) and hazy-images (HI).}\label{Resultdetection}
\begin{tabular*}{\tblwidth}{@{} LLLLLLLLLLLLLLLLL@{} }
\toprule
Method & mAP & Plane & BD & Bridge & GTF & SV & LV & Ship & TC & BC & ST & SBF & RA & Harbor & SP & HC\\
\midrule
HFI(Our) & 0.591 & 0.905 & 0.805 & 0.124 & 0.514 & 0.415 & 0.453 & 0.420 & 0.897 & 0.633 & 0.758 & 0.729 & 0.687 & 0.568 & 0.448 & 0.511\\
HI & 0.483 & 0.901 & 0.775 & 0.117 & 0.445 & 0.302 & 0.261 & 0.380 & 0.900 & 0.448 & 0.662 & 0.472 & 0.588 & 0.280 & 0.298 & 0.413\\
\bottomrule
\end{tabular*}
\begin{tablenotes}
\item[1] The short name for each category can be found in \autoref{data}.
\end{tablenotes}
\end{threeparttable}
\end{table*}

\begin{figure*}
	\centering
		\includegraphics[]{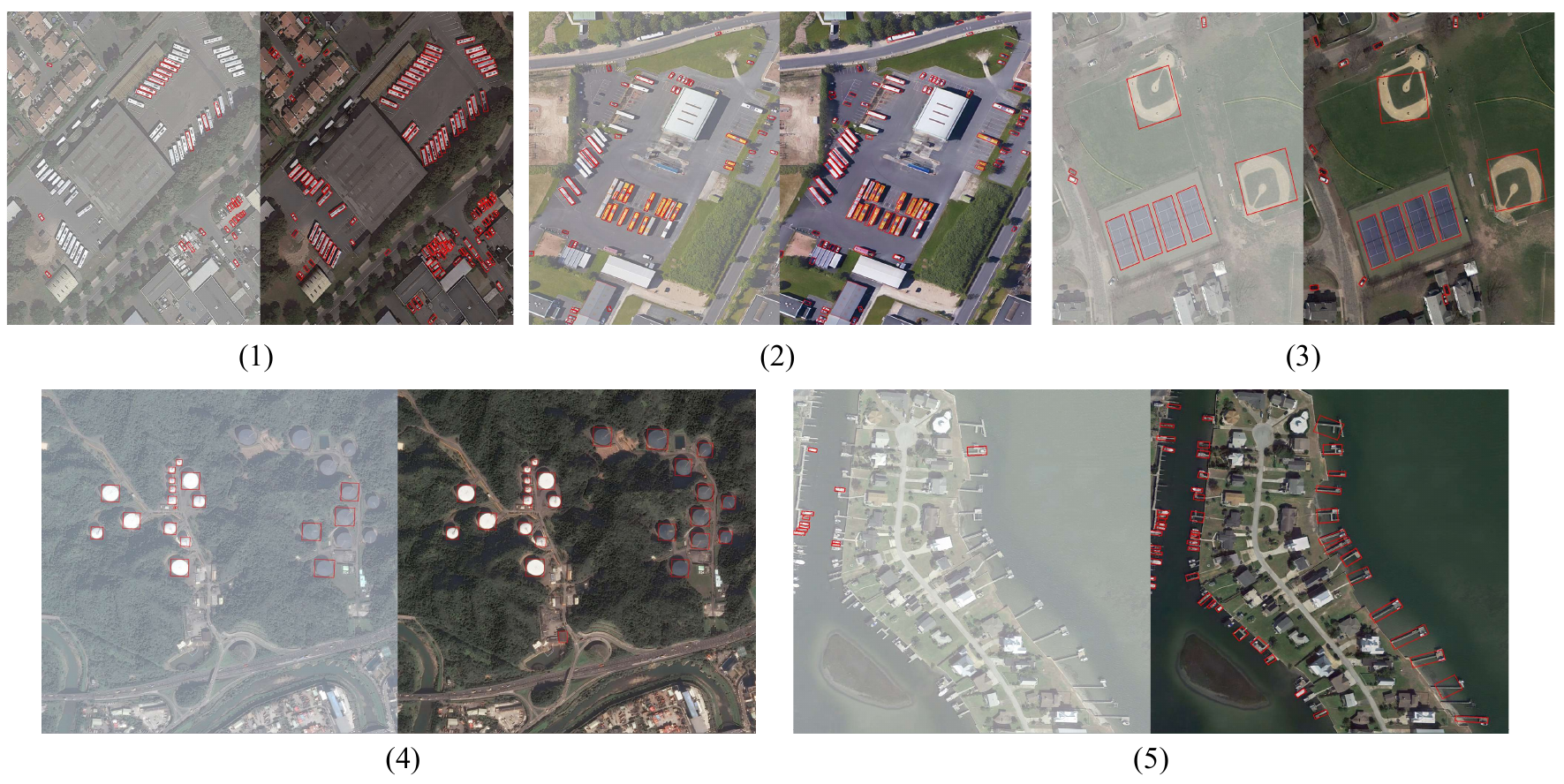}
    
	\caption{The results of object detection on Test Dataset E based on Faster-RCNN. {The red polygons are the detected object of 15 categories, which can be modified by reannotating the data and retraining the algorithm according to the demand of researches during environmental monitoring.}}
	\label{detection}
\end{figure*}

\subsection{Object detection results on Test Dataset E}
To illustrate the application of our method in environmental monitoring, we train a detection network based on Faster-RCNN\citep{Ren_2017} using DOTA. The object detection results are shown in \autoref{Resultdetection}, as can be seen in the table, the hazy-free images using proposed method (HFI) achieve the mAP of 0.591, it outperforms the hazy images (HI) by 0.108 points. In general, almost all categories have improved with our dehazing algorithm. Besides, there is a significant improvement in densely packed small instances, such as small vehicles, large vehicles and harbor. For example, the detection performance for the harbor category gains an improvement of 0.288 points compared to the results of hazy images. We give some qualitative comparison results of object detection in \autoref{detection}. in the figure, on the left of each pair is the hazy image and the right is the hazy-free image. And objects of interest are framed by red polygons, which can be modified according to the demand during environmental monitoring. In this experiment, the objects of interest are 15 categories mentioned in \autoref{data}. From the figure, we can clearly see that our method improves the detection accuracy, especially in some categories, such as small vehicles large vehicles and harbor, which is also consistent with the mAP in \autoref{Resultdetection}. Overall, our model shows the ability to improve detection results in remote sensing images, including less missing detections and inaccurate localizations. {In addition, the detection process includes object-based classification and regression, therefore, our model can also be applied to improve the accuracy of object-based classification in remote sensing images. In the following work, we will further verify the application of the proposed algorithm in classification and semantic segmentation tasks.}

\section{Conclusion}
This study proposes a feature-supervised generative adversarial network model for environmental monitoring in hazy days. The impacts of the loss function, discriminator and the similarity of the paired images during the encoding process on the model are fully considered. Furthermore, through comparison experiments with other state-of-the-art models on the same synthetic dataset, it was found that the proposed model has higher performance taking the PSNR and SSIM as indicators. The main conclusions of this study are as follows.

(1)The addition of feature information from paired clean images, which contributes to the training of encoder where hazy images are put into, can considerably improve the performance of the generator.

(2)The application of several loss functions displays good results. The model with feature regularization loss (${L}_{FR}$)  and style loss (${L}_{S}$) was found to improve the PSNR and SSIM of the basic model up to 2\% and 1\% in the verification stage, respectively.

(3)The results of dehazing and object detection on remote sensing images show the proposed model can remove haze not only in remote sensing images but also in images obtained by UAV and ground camera and improve environmental detection results, including less missing detections and inaccurate localizations.

{However, limitations remain in our study. For example, due to the limited computing capability of  GPU, it is not capable of processing for the large size of images. In this case, we consider cropping the image first and then merging them into the original size after processing. In addition, our model uses RGB images only in remote sensing images. there also have important information in other non-visible bands. To improve the performance of the proposed algorithm, other bands and band combinations should be considered and examined.}

\section*{Acknowledgment}
This research was funded by National Natural Science Foundation of China (51605054), State Key Laboratory of Vehicle NVH and Safety Technology (NVHSKL-202008, NVHSKL-202010), The Science and Technology Research Program of Chongqing Education Commission of China (NKJQN201800517 and KJQN201800107), Fundamental Research Funds for the Central Universities (2019CDXYQC003), Chongqing Social Science Planning Project (No:2018QNJJ16)

\bibliographystyle{cas-model2-names}

\bibliography{MyPaperReference}


\end{document}